\pdfoutput=1
\documentclass[a4paper,11pt]{article}
\usepackage[utf8]{inputenc}
\usepackage{geometry}
\usepackage{amsmath}
\usepackage{amssymb}
\usepackage{graphics}
\usepackage[pdftex]{graphicx, color}
\usepackage{pdfpages}
\usepackage{relsize}
\usepackage{natbib}
\usepackage{siunitx}
\usepackage{authblk}
\begin{document}
\title{\Large \bf Forest-based methods and ensemble model output statistics for rainfall ensemble forecasting}
\author[1,2,3]{Maxime Taillardat}
\author[2]{Anne-Laure Foug\`eres}

\author[3]{Philippe Naveau}

\author[1]{Olivier Mestre}
\affil[ ]{}
\affil[1]{CNRM UMR 3589 and DirOP/COMPAS, M\'et\'eo-France/CNRS, Toulouse, France}
\affil[ ]{}
\affil[2]{Univ Lyon, Universit\'e Claude Bernard Lyon 1, CNRS UMR 5208, Institut Camille Jordan, 43 blvd. du 11 novembre 1918, F-69622 Villeurbanne cedex, France}
\affil[ ]{}
\affil[3]{Laboratoire des Sciences du Climat et de l'Environnement (LSCE-CNRS-CEA-UVSQ-IPSL), Gif-sur-Yvette, France}

\date{}
\affil[ ]{}
\affil[ ]{}
\affil[ ]{}
\affil[ ]{maxime.taillardat@meteo.fr}
\maketitle

\begin{abstract}
Rainfall ensemble forecasts have to be skillful for both low precipitation and extreme events. 
We present statistical post-processing methods based on  Quantile Regression
Forests (QRF) and Gradient Forests (GF) with a parametric extension for heavy-tailed distributions.
Our goal is to improve ensemble quality for all types of precipitation events, heavy-tailed included, subject to a good
overall performance.

Our hybrid proposed methods are applied to daily 51-h forecasts of 6-h accumulated precipitation from 2012 to 2015 over
France using the M\'et\'eo-France ensemble prediction system called PEARP. 
They provide calibrated predictive
distributions and compete favourably with state-of-the-art methods like Analogs method or Ensemble Model Output Statistics. 
In particular, hybrid forest-based procedures appear to bring an added value to the forecast of heavy rainfall.

\end{abstract}

\section{Introduction}
\subsection{Post-processing of ensemble forecasts}
Accurately forecasting   weather is paramount for a wide range of end-users, e.g.   air traffic controllers,   emergency managers and energy providers \citep[see, e.g.][]{pinson2007trading, zamo2014benchmark}. 
In meteorology, ensemble forecasts try to quantify forecast uncertainties  due to  observation errors and incomplete
physical representation of the atmosphere.   
Despite its recent developments in national meteorological services, ensemble forecasts still  suffer of bias and underdispersion \citep[see, e.g.][]{hamill1997verification}. 
Consequently, they need to be  post-processed. 
At least two types of statistical methods have emerged in the last decades:   
 analogs method   and  ensemble model output statistics (EMOS) \citep[see, e.g.][respectively]{delle2013probabilistic,gneiting2005calibrated}.
 The first one is fully non-parametric and consists in finding similar atmospheric situations in the past  and using them to improve the present forecast.
 In contrast, EMOS belongs to the family of   parametric regression schemes.
If $y$ represents the    weather variable of interest and $(x_1,\dots, x_m)$   the corresponding $m$ ensemble member forecasts, then 
  the EMOS predictive distribution is simply  a  distribution whose  parameters 
   depend on the values of  $(x_1,\dots, x_m)$. 
 Less conventional approaches 
have also  been  studied recently. For example,    \citet{van2015ensemble} investigated   member-by-member post-processing techniques 
and 
 \citet{taillardat2016calibrated}  found that  quantile regression forests (QRF) techniques performed well for  temperatures and wind speed data.

\subsection{Forecasting and calibration of precipitation}

Not all meteorological variables are equal in terms of forecast and calibration.  
In particular,  \citet{hemri2014trends} highlighted  that rainfall forecasting  represents a  steep hill. 
In this study, we  will  focus on 6-h rainfall amounts in France because this is the unit of interest of the 
ensemble forecast system of M\'et\'eo-France. 
For daily  precipitation,  
%
%
extended logistic regression was frequently applied \citep[see, e.g.][]{hamill2008probabilistic,roulin2012postprocessing,ben2013calibrated}. 
Bayesian Model Averaging  techniques \citep{raftery2005using,sloughter2007probabilistic}  were also used in rainfall  forecasting, but we will not cover them here 
 because a  gamma fit is often applied  to cube root transformed precipitation accumulations   and this complex transformation may not  be adapted to 6h rainfall.  
Concerning analogs and  EMOS   techniques, they  
have been applied     to calibrate    daily rainfall \citep[see][]{hamill2006probabilistic,scheuerer2014probabilistic,scheuerer2015statistical}. 
As the QRF method in  \citet{taillardat2016calibrated} performed  better than EMOS   for temperatures and wind speeds, one may wonder if 
QRF could favourably compete with  EMOS and analogs  techniques for rainfall calibration. 
This question is particularly relevant because  recent methodological advances have been made concerning random forests and quantile regressions. 
In particular,  \citet{athey2016solving}   proposed an innovative 
   way,  called gradient forests (GF),  of using forests  to make quantile regression.  
In this context, we propose to implement and  test this quantile regression GF method for rainfall calibration and compare it with other approaches, see Section \ref{sec: QRF and GF}.

\subsection{Parametric probability density functions (pdf) of precipitation}\label{sec: 3 pdfs}
Modeling precipitation distributions is a challenge by itself. 
It is a mixture of zeros (dry events) and positive intensities, i.e. rainfall amounts for wet events. 
The latter have a skewed distribution. 
One popular and flexible choice to model rainfall amounts is to use the gamma distribution or to built on it. 
For example, \citet{scheuerer2015statistical} and \citet{baran2016censored} 
 in a rainfall calibration context employed the  censored-shifted gamma (CSG) pdf defined by 
 \begin{eqnarray}\label{eq: CSG}
 f_{CSG}(y) &=& \left\{ \begin{array}{ll}
   (1-\pi)  \cdot  \frac{(y+\delta)^{\kappa-1}}{\Gamma(\kappa)}  \exp(- (y+\delta)/\theta), & \; \;  \mbox{if } y>0 \\
  \pi, & \; \; \mbox{if } y=0,
  \end{array}                 
             \right.
\end{eqnarray}
where $y\geq 0$, the positive constants  $(\kappa,\theta)$ are the two gamma law parameters  and  the probability $\pi \in [0,1]$ represents the mass of the gamma cumulative distribution function (cdf)   below  the level of censoring $\delta\geq 0$. 
Hence, the probability of zero and positive  precipitation are treated together.
One possible drawback of the CSG is that  heavy daily and subdaily rainfall may not always have a nice upper  tail with an exponential decay like a gamma distribution, but rather 
a polynomial one, the latter point being 
a  key element in any weather risk analysis \cite[see, e.g.][]{katz2002statistics,de2007extreme}.
To bring the necessary flexibility in modelling    upper tail behavior in a rainfall EMOS context,  
 \citet{scheuerer2014probabilistic} worked with a so-called censored generalized extreme value (CGEV) defined by 
  \begin{eqnarray}\label{eq: CGEV}
 f_{CGEV}(y) &=& \left\{ \begin{array}{ll}
   (1-\pi)  \cdot g(y;\mu,\sigma,\xi), & \; \;  \mbox{if } y>0 \\
  \pi, & \; \; \mbox{if } y=0,
  \end{array}                 
             \right.
\end{eqnarray}
where $\pi = G(0;\mu,\sigma,\xi)$ and  the pdf $g(y;\mu,\sigma,\xi)$ which cumulative distribution function $G$ is the classical GEV 
$$
G(y;\mu,\sigma,\xi) =   \exp\left[-\left(1+\frac{\xi(y-\mu)}{\sigma}\right)_+^{-1/\xi}\right]  \mbox{ for $\xi\neq0$.}
$$
Note that $a_+=\max(0,a)$ and that, if $\xi =0$, then  $g(y;\mu,\sigma,0)$ represents the classical Gumbel pdf. 
To be in compliance with extreme value theory (EVT) not only for heavy rainfall but also for low precipitation
amounts,  
  \citet{WRCR:WRCR21974} recently proposed a   class of models  
 referred as the extended generalized
Pareto (EGP)  that allows a smooth transition between generalized Pareto (GP) type 
tails and the middle part (bulk) of the distribution.   
It  bypasses the complex thresholds selection step to define extremes. 
Low precipitation can be shown to be  gamma distributed, while heavy rainfall are Pareto distributed. 
Mathematically, a cdf belonging to the    EGP family has to be expressed as 
\begin{equation*}
T\left\{ H_{\xi}(y/\sigma)\right\} \mbox{, for all $y>0$,}
\end{equation*}
 where 
  $H_{\xi}(y)=1- \left( 1+ \xi y\right)^{-1/\xi} $ represents the  GP  cdf, while $T$ denotes a continuous cdf on
the unit interval. 
To insure that the upper   tail behavior of $T$ is driven by the shape parameter $\xi$, the survival function  $\bar{T}=1-T$ has to satisfy  that  
$\underset{u\downarrow0}{\lim}\frac{\bar{T}(1-u)}{u}$   is finite. 
To force low rainfall to follow a GPD for small values near zero, we need that $\underset{u\downarrow0}{\lim}\frac{T(u)}{u^{s}}$   is finite for some real $s>0$. 
Studies have already made this choice \cite[see, e.g.][]{vrac2007stochastic,WRCR:WRCR21974}. 
In \citet{WRCR:WRCR21974}, different  parametric models  of the cdf  $T$     satisfying  the required constraints     were compared. 
The special case where $T(u)=u^{\kappa}$ with $\kappa>0$ obeys these constraints and also  corresponds to a model studied by \citet{papastathopoulos2013extended}.
In practice,  this simple version of $T$ appears to fit well daily and subdaily rainfall and consequently, we will only focus on this case in this paper. 
In other words, our third model for the precipitation pdf  is 
  \begin{eqnarray}\label{eq: EGPD}
 f_{EGP}(y) &=& \left\{ \begin{array}{ll}
   (1-\pi) \cdot  \frac{\kappa}{\sigma} \cdot \left\{ H_{\xi}(x/\sigma)\right\}^{\kappa-1} \cdot h_{\xi}(y/\sigma), & \; \;  \mbox{if } y>0 \\
  \pi, & \; \; \mbox{if } y=0,
  \end{array}                 
             \right.
\end{eqnarray}
where $h_{\xi}(.)$ is the pdf associated with $H_{\xi}(.)$.
In   contrast to (\ref{eq: CSG}) and  (\ref{eq: CGEV}), 
the probability weight $\pi$ is not obtained by censoring, and it is just a parameter independent of $(\kappa,\sigma,\xi)^T$.

At this stage, we have three parametric pdfs, see (\ref{eq: CSG}) and  (\ref{eq: CGEV}) and   (\ref{eq: EGPD}), to implement a EMOS approach to 6-hour rainfall data, see Section \ref{sec: EMOS}. 
Besides comparing these three EMOS models, 
it is  natural to wonder if QRF and GF methods could take advantage of these three parametric forms.

\subsection{Coupling parametric pdfs with  random forest approaches}\label{sec:  coupling}
 A drawback of data driven approaches like QRF and GF   is that their intrinsic  non parametric nature make  them useless to   predict beyond the largest recorded rainfall. 
To circumvent  this limit, we also propose to combine  random forest techniques with a EGP pdf defined by  (\ref{eq: EGPD}), see Section \ref{sec: hybrid}. 
Hence,  random forest-based post-processing techniques  will be in compliance with EVT and this should be   an interesting path to improve prediction behind the largest values of the sample at hand.

  \subsection{Outline}
  This article is organized as follows.
  In Section \ref{sec: QRF and GF}, we recall the basic ingredients to create quantile regression forests and gradient forests. 
  In particular,  we review   the calibration process of the GF method recently introduced by \citet{athey2016solving} for quantile regression.
  Then, we explain how these trees are  combined with the EGP pdf  defined by  (\ref{eq: EGPD}).

    In Section \ref{sec: EMOS}, we propose to integrate the  EGP pdf within a EMOS scheme. 
    
    The different approaches are implemented in Section \ref{sec: PEARP} where the test bed dataset of   87 French weather stations  and the French
ensemble forecast system of M\'et\'eo-France called PEARP \citep{descamps2014pearp} is described.  
Then, we assess and compare each method with a special interest for heavy rainfall, see Section \ref{sec: results}. 
The paper closes with a discussion in Section \ref{sec: end}.

\section{Quantile regression forests and gradient forests}\label{sec: QRF and GF}

\subsection{Quantile regression forests}
Given a sample of predictors-response pairs, say $(X_i,Y_i)$ for $i=1,\dots,n$, classical regression techniques  connect  the conditional mean of a response variable $Y$  to a given set of predictors $X$. 
The quantile regression forest (QRF) method  introduced by \citet{meinshausen2006quantile} also consists in building a link, but between  an empirical cdf and  
the outputs of a tree.
Before explaining this particular cdf, we need to recall how trees are constructed.

A random forest is an aggregation of randomized trees based on  bootstrap aggregation  on the one hand, and on classification and regression trees (CART) 
\citep{breiman1996bagging,breiman1984classification} on the other hand.  These trees are built on a bootstrap copy of the samples by recursively maximizing a splitting rule.
 Let $\mathcal{D}_0$ denote the group of observations to be  divided into  two subgroups, say  $\mathcal{D}_1$ and $\mathcal{D}_2$.  
For each group, we can infer its  homogeneity defined by  
\begin{equation*}
 v(\mathcal{D}_j)=\sum_{Y \in \mathcal{D}_j}[Y-\overline{Y}(\mathcal{D}_j)]^2,
\end{equation*}
 where $\overline{Y}(\mathcal{D}_j)$ corresponds to the sample mean   in $\mathcal{D}_j$. 
 To determine if this splitting choice is optimal, the homogeneities $ v(\mathcal{D}_1)$ and $ v(\mathcal{D}_2)$ are compared to the one of $\mathcal{D}_0$.
 For example, if wind speed is one predictor in $X$ and     dividing   low and large winds could better explain rainfall, then 
 the cutting value, say $s$,  will be the one that maximizes
 \begin{equation}
 \mathcal{H}(\mathcal{D}_1,\mathcal{D}_2)=\underset{s \in \mathcal{E}^*}{\mathrm{max}}\left[ v(\mathcal{D}_0)-v(\mathcal{D}_1)-v(\mathcal{D}_2)\right]
 \end{equation}
 where  $\mathcal{E}^*$ is a random subset of the predictors  in the predictors' space $\mathcal{E}$. 
 Each resulting group is itself split into two, and so on until some stopping criterion is reached.
As each tree is built on a random subset of the predictors, the method is called   ``random forest" \citep{breiman2001random}.
Binary regression trees can be viewed as    decision trees, each node being the criterion used to split the data and each final leaf giving the predicted value. 
For example, if we observe a given wind speed $x$, we  can find the final leaf that corresponds to this value of $x$ and the associated observations $y$, then we can compute 
 the conditional cumulative distribution function   introduced by \citet{meinshausen2006quantile}
\begin{equation}\label{eq: ecdf}
 \widehat{F}(y|x) = \sum_{i=1}^{n} \omega_i(x)\mathbf{1}(\{ Y_i \leq y\}) ,
\end{equation}
where the weights $\omega_i(x)$ are deduced from the presence of $Y_i$ in a final leaf of each tree when one follows the path determined by $x$.
The interested reader is referred  to \citet{taillardat2016calibrated} for an application of this  approach to  ensemble forecast  of temperatures and winds.

\subsection{Gradient forests}
 \citet{meinshausen2006quantile} proposed splitting rule using CART regression splits. 
 Arguing that this
 splitting rule is not tailored to the quantile regression context, \citet{athey2016solving} proposed another optimisation  scheme. Instead of maximizing the variance heterogeneity of the children nodes, one maximizes
 the criterion 
  \begin{equation}
  \Delta(\mathcal{D}_1,\mathcal{D}_2)=\sum_{j=1}^2 \frac{-1}{|\{i : Y_i \in \mathcal{D}_j\}|}{\left(\sum_{\{i : Y_i \in \mathcal{D}_j\}} \rho_i\right)}^2 \end{equation} where
  the indicator function 
$
 \rho_i=\mathbf{1}(\{Y_i\geq \hat\theta_{q,\mathcal{D}_0}\}) 
$ 
is equal to one when  $Y_i$ is greater than  the q-th quantile $\hat\theta_{q,\mathcal{D}_0}$ of the observations of the parent node $\mathcal{D}_0$. 
 The terminology of \textit{gradient forests} was suggested  because the choice of $\rho_i$ is here linked with a gradient-based approximation of the quantile function 
\begin{equation*}
 \Psi_{\hat\theta_{q,\mathcal{D}_0}}(Y_i)=q\mathbf{1}(\{Y_i > q\})+(1-q)\mathbf{1}(\{Y_i \leq q\}).
\end{equation*}
This technique using gradients is computationally feasible, an issue not to be omitted when dealing with non-parametric techniques. Note here that for each split the order of the quantile is chosen among given orders 
$(0.1,0.5,0.9)$.
In the special case of least-square regression,   $\rho_i$ becomes $Y_i-\overline{Y}(\mathcal{D}_0)$, and 
  $\mathcal{H}(\mathcal{D}_1,\mathcal{D}_2)$ becomes equivalent to $\Delta(\mathcal{D}_1,\mathcal{D}_2)$. In this special case,  gradient trees are equivalent to build a standard CART regression tree.

\subsection{Fitting a parametric form to  QRF and GF trees}\label{sec: hybrid}
As mentioned in Section \ref{sec:  coupling},  the predicted cdf  defined by (\ref{eq: ecdf}) cannot predict values which are not in the learning sample.
This can be a strong limitation if the learning sample sample is small or rare events are of interest or both. 
The GF method has the same issue.
To parametrically model rainfall, the EGP pdf  defined by (\ref{eq: EGPD}) appears to be a good candidate. It allows more flexibility in the fitting than CSG or CGEV.
This distribution has four parameters, $\pi, \kappa, \sigma$ and $\xi$, it is in compliance with EVT for low and heavy rainfalls and works well in practice \cite[see, e.g.][]{WRCR:WRCR21974}.
In terms of inference, a simple and fast method-of-moment can be applied.
Basically, probability weighted moments (PWM) of a given  random variable, say $Y$, with survival function $\overline{F}(y)=\mathbf{P}(Y>y)$, can be expressed as \cite[see, e.g.][]{hosking1987parameter} 
\begin{equation}\label{eq: pwm}
{\mu}_r=\mathbf{E}([Y \overline{F}^r(Y)])= \int_0^1 F^{-1}(q) (1-q)^r dq.
\end{equation}
If $Y$ follows a EGP pdf defined by  (\ref{eq: EGPD}), then we have 
\begin{eqnarray*}
\frac{\xi} {\sigma} {\mu}_0 &=&  \kappa B(\kappa, 1-\xi) -1  \mbox{ and } 
\frac{\xi} {\sigma} {\mu}_1 =  \kappa \left(B(\kappa, 1-\xi) - B(2\kappa, 1-\xi)\right)-\frac{1}{2}, \\
\frac{\xi} {\sigma} {\mu}_2& =&  \kappa \left(B(\kappa, 1-\xi) - 2B(2\kappa, 1-\xi) + B(3\kappa, 1-\xi)\right)-\frac{1}{3}, 
\end{eqnarray*}
where $B(.,.)$ represents the beta function. 
Knowing the  PWM triplet $( {\mu}_0, {\mu}_1,  {\mu}_2)^T$ is equivalent to know the parameter vector $(\kappa,\sigma,\xi)^T$.
Hence, we just need to estimate these three PWMs. 
For any given  forest, it is possible to estimate the distribution of $[Y|X=x]$ by the empirical cdf  $\widehat{F}(y|X=x)$  defined by (\ref{eq: ecdf}). 
Then, we can plug it in (\ref{eq: pwm}) to get 
$$
\hat{{\mu}}_r(x)=\  \int_0^1 \widehat{F}^{-1}(q|X=x) (1-q)^r dq.
$$
This leads to the estimates of $(\kappa(x),\sigma(x),\xi(x))^T$ and consequently of $f(y |X=x)$ via Equation (\ref{eq: EGPD}). 
Note that the probability of no rain $\pi(x)$ is just inferred by counting the number of dry events in the corresponding trees. 
In the following, this technique is called "EGP TAIL", despite the fact that the whole distribution is fitted from QRF and GF trees.

\section{Ensemble model output statistics and EGP}\label{sec: EMOS}
In Section \ref{sec: 3 pdfs}, three definitions of  parametric pdfs were recalled. 
By regressing their parameters   on the ensemble values, different EMOS models have been proposed for the CSG and  CGEV pdfs defined by (\ref{eq: CSG})  and  by (\ref{eq: CGEV}), respectively. 
More precisely, 
\citet{baran2016censored} used the CSG pdf   
by  letting the mean $\mu = \kappa \theta$ and variance   
 $\sigma^2=\kappa \theta^2$ depend linearly  as functions  of the raw ensemble values and their mean, respectively. 
The coefficients of this regression were   estimated by  miminizing the continuous ranked probability score (CRPS) \cite[see, e.g.][]{scheuerer2015statistical, hersbach2000decomposition}.   
The same strategy can be applied to fit the CGEV pdf \cite[see, e.g.][]{hemri2014trends}. \citet{scheuerer2014probabilistic} modelled the scale parameter  $\sigma$  in (\ref{eq: CGEV})
 as an affine function of the ensemble mean absolute deviation rather than of  the raw ensemble mean or variance.  
 Another point to emphasize is that the shape parameter $\xi$ was considered invariant in space in \citet{hemri2014trends}.   

\begin{figure}[!h]
  \includegraphics[width=\textwidth,angle=0]{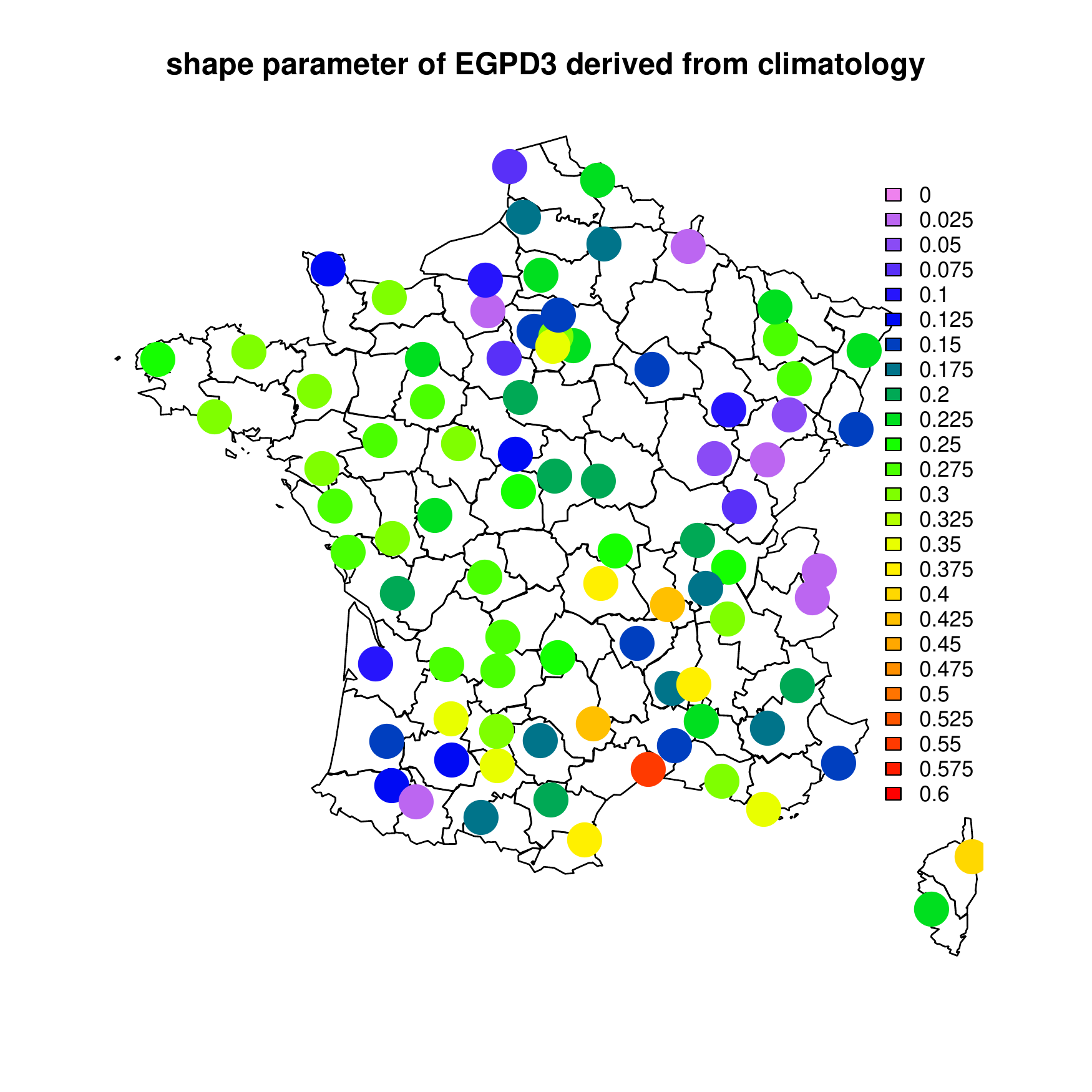}\\
  \caption{Spatial values of  $\xi$ among locations.}\label{climxi}
\end{figure}

In this section, 
we basically explain how  an EMOS approach can be built with the EGP pdf defined by (\ref{eq: EGPD}) 
and we  now highlight common features and differences  between the two EMOS with CSG and CGEV. The scale parameter $\sigma^2$ in (\ref{eq: EGPD})  is estimated in the same way than for CGEV.
 The presence of 
the parameter $\kappa$ allows an additional degree of freedom. The expectation of our EGP is mainly driven by the product $\kappa\sigma$. 
Consequently, we model $\kappa$ as  an affine function of the predictors divided by $\sigma$.
As France has a diverse climate, it is not reasonable to assume a constant shape parameter among all  locations, see the map in Figure \ref{climxi}.
In addition, minimizing  the CRPS to infer different shape parameters may be inefficient  \cite[see, e.g.][]{friederichs2012forecast}. 
To estimate $\xi$ at each location, we simply use the PWM inference scheme described in Section \ref{sec: hybrid}. To complete the estimation of the parameters in (\ref{eq: EGPD}), the probability $\pi$ is modeled as an affine function on $[0,1]$ of the raw ensemble probability of rain
and affine function parameters are also estimated by CRPS minimization. The table \ref{esti} sums up the optimal estimation strategies that we have found for each distribution.

\begin{table*}[!h]
\caption{ Optimal strategies for parameter estimation using CRPS minimization in the EMOS context.}\label{esti}
\begin{center}
\small
\begin{tabular}{ccc}
\hline
\hline
 Distribution &  Parameter & Comments\\
\hline

  CSG & $\delta$ & free in $\mathbb{R}$ \\
  &$\mu$& affine function of covariates in C\\
  &$\sigma$&affine function of raw ensemble mean\\
  &$\kappa$& $\kappa=\mu^2/\sigma$\\
  &$\theta$& $\theta=\sigma/\mu$\\
 \hline
   CGEV & $\mu$& affine function of covariates in C\\
  &$\sigma$&affine function of the mean absolute deviation of the raw ensemble \\
  &$\xi$& free in $(-\infty,1)$\\
  &$\theta$& $\theta=\sigma/\mu$\\
 \hline
   EGP & $\sigma$&affine function of the mean absolute deviation of the raw ensemble \\
  &$\mu$& maximum between 0 and an affine function of covariates in C \\
  &$\kappa$&$\kappa=\mu/\sigma$\\
  &$\xi$& fixed, see Figure \ref{climxi} for stations' values\\
  &$\pi$& affine function of PR0 in C, bounded on $[0,1]$\\
\hline
\end{tabular}
\normalsize
\end{center}
\end{table*}

\section{Case study on the PEARP ensemble prediction system}\label{sec: PEARP}

\subsection{Data description}
Our rainfall dataset  corresponds to 6-h rainfall amounts produced by  87 French weather stations and the 35-member ensemble forecast system called PEARP \citep{descamps2014pearp}  at     a 51-h lead time forecast.
Our period of interest spans    four years from  1 January 2012 to 31 December 2015.

\subsection{Inferential details for EMOS and analogs}
Verification has been made on this entire period. For a fair comparison each method has to be tuned optimally. EMOS uses all the data available for each day (4 years less the forecast day as a training period).  
 The same strategy is used to fit the analogs method, see  Appendix \ref{sec: analogs} for details on this approach. 
 QRF and GF employ a cross-validation method: each month of the 4 years is kept as validation data while
  the rest of the 4 years is used for learning. The tuning algorithm for EMOS is stopped after few iterations in order to avoid overfitting, as suggested in \citet{scheuerer2014probabilistic} concerning the parameter estimations.

\subsection{Sets of predictors used}
We either use a subset of classical predictors (denoted by ``C'' in the rest of the paper) detailed in Table \ref{setc} or the whole set of available predictors as listed in Table \ref{seta}.

  \begin{table*}[!h]
\caption{Subset ``C'' representing the most classical predictors.}\label{setc}
\begin{center}
\small
\begin{tabular}{cc}
\hline
\hline
Name & Description\\
\hline
HRES & high resolution member \\
CTRL & control member \\
MEAN & mean of raw ensemble \\
PR0 & raw probability of rain \\
\hline
\end{tabular}
\normalsize
\end{center}
\end{table*}

\begin{table*}[!h]
 \caption{Set of all available predictors.}\label{seta}
\begin{center}
\scriptsize
\begin{tabular}{cc}
\hline
\hline
Name & Description\\
\hline
HRES  &\multicolumn{1}{c}{high resolution member}\\
CTRL  &\multicolumn{1}{c}{control member}\\
MEAN  &\multicolumn{1}{c}{mean of raw ensemble}\\
MED  &\multicolumn{1}{c}{median of raw ensemble}\\
Q10  &\multicolumn{1}{c}{first decile of raw ensemble}\\
Q90  &\multicolumn{1}{c}{ninth decile of raw ensemble}\\
PR0  &\multicolumn{1}{c}{raw probability of rain}\\
PR1  &\multicolumn{1}{c}{raw probability of rain $>1$mm/6h}\\
PR3  &\multicolumn{1}{c}{raw probability of rain $>3$mm/6h}\\
PR5  &\multicolumn{1}{c}{raw probability of rain $>5$mm/6h}\\
PR10  &\multicolumn{1}{c}{raw probability of rain $>10$mm/6h}\\
PR20  &\multicolumn{1}{c}{raw probability of rain $>20$mm/6h}\\
SIGMA  & standard deviation of raw ensemble \\
IQR  & IQR of raw ensemble \\
HU1500  & deterministic forecast of 6-h mean 1500m humidity  \\
UX  & deterministic forecast of 6-h maximum of zonal wind gust \\
VX  & deterministic forecast of 6-h maximum of meridional wind gust \\
FX  & deterministic forecast of 6-h maximum of wind gust power\\
TCC  & deterministic forecast of 6-h mean total cloud cover\\
RR6CV  & deterministic forecast of 6-h convective rainfall amount\\
CAPE  & deterministic forecast of 6-h mean convective available potential energy \\
\multicolumn{2}{c}{}\\
\multicolumn{2}{c}{q10,50,90 are the first decile, the median and ninth decile of the raw ensemble for these variables:}\\
\multicolumn{2}{c}{}\\
HU\_q10,50,90  &\multicolumn{1}{c}{6-h mean surface humidity} \\
P\_q10,50,90 &\multicolumn{1}{c}{6-h mean sea level pressure} \\
TCC\_q10,50,90  &\multicolumn{1}{c}{6-h mean total cloud cover} \\
RR6CV\_q10,50,90  &\multicolumn{1}{c}{6-h convective rainfall amount} \\
U10\_q10,50,90  &\multicolumn{1}{c}{6-h mean surface zonal wind} \\
V10\_q10,50,90  &\multicolumn{1}{c}{6-h mean surface meridional wind} \\
U500\_q10,50,90  &\multicolumn{1}{c}{6-h mean 500m zonal wind} \\
V500\_q10,50,90  &\multicolumn{1}{c}{6-h mean 500m meridional wind} \\
FF500\_q10,50,90  &\multicolumn{1}{c}{6-h mean 500m wind speed} \\
TPW850\_q10,50,90  &\multicolumn{1}{c}{6-h mean 850hPa potential wet-bulb temperature} \\
FLIR6\_q10,50,90  &\multicolumn{1}{c}{6-h mean surface irradiation in infra-red wavelengths} \\
FLVIS6\_q10,50,90  &\multicolumn{1}{c}{6-h mean surface irradiation in visible wavelengths} \\
T\_q10,50,90 & 6-h mean surface temperature \\
FF10\_q10,50,90 & 6-h mean surface wind speed \\
\hline
\end{tabular}
\normalsize
\end{center}
\end{table*}

 Note that we also considered for EMOS a third type of predictors set  based on a variable selection algorithm (see Appendix \ref{app3}). 
But this did not improve the results and we removed them from the analysis (available upon request).

\subsection{Zooming on extremes}\label{sec: zoom}
Finding a way to assess the quality of ensembles for extreme and rare events is quite difficult, as seen in \citet{williams2014comparison} in a comparison of  ensemble calibration methods for extreme events.  
Weighted scoring rules can be adopted as done in \citet{gneiting2011comparing, lerch2017forecaster} but there are here two main issues. The ranking of compared methods depends on the weight function used, as already suggested in \citet{gneiting2011comparing}. Besides, giving a weight to such rare events avoid discriminant power of scoring rules, the same issue than for the Brier score \citep{brier1950verification}. Moreover, reliability is not sound here since there 
are not enough extreme cases (by definition) 
to measure it. We have finally decided to focus on two ideas here, matching with forecasters' desires: first, what is the discriminant power of our forecasts for extreme events in terms of binary decisions ? Second, 
what is the potential risk of our ensemble to mismatch an extreme event ?
The choice done in our study is discussed in Section \ref{sec: results}.

\section{Results}\label{sec: results}

\begin{figure}[!h]
  \includegraphics[width=\textwidth,angle=0]{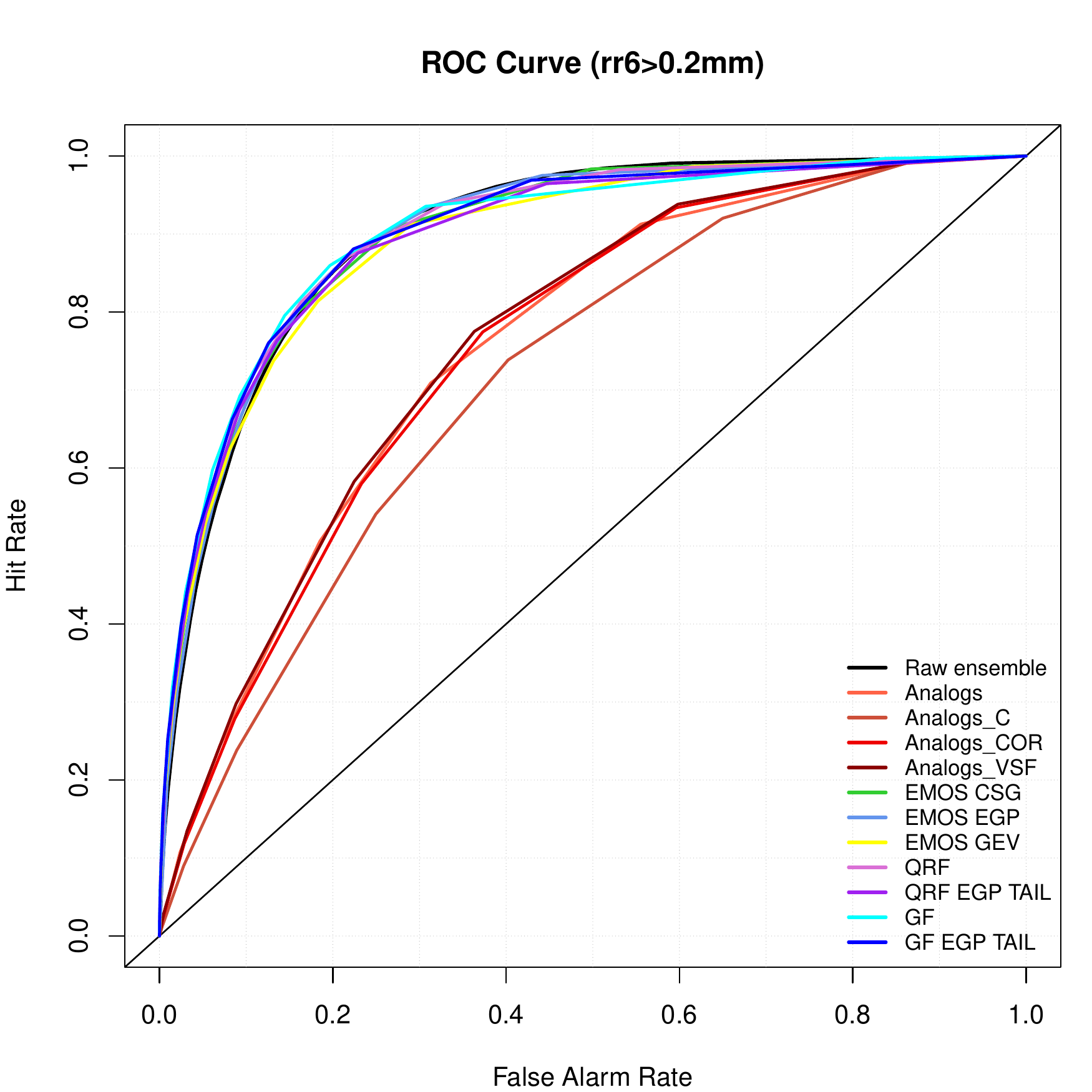}\\
  \caption{ROC Curves for the event of rain. A ``good'' prediction must maximize hit rate and minimize false alarms. The analogs method lacks resolution. We 
  can notice that there is no improvement of post-processed methods compared to the raw ensemble.}\label{roc}
\end{figure}

Table \ref{res} compares different metrics for all post-processing techniques which have been fitted to the 87 stations and averaged over  4 years of verification. Ten methods are competing: 
 The raw ensemble, 4 analogs, 
3 EMOS (3 different distributions using the set C), 2 forest-based methods (1 QRF and 1 GF) and 2 tail-extended forest-based methods (1 QRF and 1 GF).
Scores used concern respectively (i) global performance (calibration and sharpness) measured by the CRPS; (ii) reliability performance, measured by the mean, the normalized variance and the entropy of the PIT histograms, denoted by $\Omega$ in the sequel;  (iii) gain in CRPS compared to the raw ensemble, measured by the Skill of the CRPS using the raw ensemble as baseline. A brief summary about these measures is done in  \ref{app4}, where references are also provided. And the boxplots showing rank histograms are in  \ref{app5}.\\
According to  Table \ref{res}, the raw ensemble is biased and underdispersive. The EMOS post-processed ensembles share with QRF and GF a good CRPS. Moreover, we can consider them as unbiased and mostly well-dispersed. 
The tail-extended  methods get a lower CRPS, that can be explained 
by their skill for extreme events. Finally,  the four analog methods show a quite poor CRPS compared to the raw ensemble, even if they exhibit reliability. Nevertheless we can notice that a weightning of the predictors, especially with a non-linear variable selection algorithm (Analogs\_VSF), brings benefits to this method.
This phenomenon can be explained by Figure \ref{roc}, where the ROC curves are given for the event of rain. Consider a fixed threshold $s$ and the contingency table associated to the predictor $\mathbf{1}\{rr6 > s\}$.
 Recall that the ROC curve then plots the probability of detection (or hit rate) as a function of the probability of false detection (or false alarm rate).
   A ``good'' prediction must maximize hit rates and minimize false alarms \cite[see, e.g.][]{jolliffe2012forecast}. 
   Figure \ref{roc} explicitely shows the lack of resolution of the analogs technique. Incidently, we can also notice that the rain event discrimination is not improved by post-processed ensembles.

\begin{table*}[!h]
\caption{ Comparing performance statistics for different post-processing methods for 6-h rainfall forecasts in France. The mean CRPS estimations come from bootstrap replicates, the estimation error is under 
$6.1\times 10^{-3}$ for all methods.
}\label{res}
\begin{center}
\small
\begin{tabular}{cccccccc}
\hline
\hline
 Types &  Methods & pdf & CRPS   & $\mathbf{E}(Z)$ & $\mathbf{V}(Z)$ & $\Omega$ & CRPSS\\
\hline
 &  Raw ensemble & &  0.4694  & 0.4164 & 1.0612  & 0.9809 & 0\% \\
 \hline
 Non-parametric  & Analogs    &     &  0.5277  & 0.5175 & 1.0190 & 0.9956 & -12.4\% \\
 & Analogs\_C       &  & 0.5376  & 0.5050 & 1.0051 & 0.9964 & -14.5\% \\
 & Analogs\_COR       &  & 0.5276 & 0.5062 & 1.0015 & 0.9964  & -12.4\% \\
 & Analogs\_VSF       &  & 0.5247  & 0.5060 & 0.9986 & 0.9961 & -11.8\% \\
 & QRF   &  & 0.4212  & 0.5006 & 0.9995 & 0.9961 & 10.3\% \\
 & GF   &  & 0.4134  & 0.5070 & 0.9771 & 0.9957 & 11.9\% \\
\hline
Parametric  & EMOS  & CSG  &  0.4224  & 0.4992 & 1.0363 & 0.9955 & 10.0\% \\
 with & EMOS & GEV   &  0.4228  & 0.5000 & 1.0073 & 0.9961 & 9.9\% \\
covariates $\in$ C  & EMOS  & EGP  &  0.4292  & 0.4623 & 1.0723 & 0.9905 & 8.6\% \\
\hline
Hybrid & QRF & EGP TAIL   &  0.4138  & 0.5095 & 0.9558 & 0.9957 & 11.8\% \\
& GF & EGP TAIL   &  0.4127  & 0.5152 & 0.9425 & 0.9948 & 12.1\% \\
\hline
\end{tabular}
\normalsize
\end{center}
\end{table*}

To sum up,  the best improvement with respect to  the raw ensemble is for the  forest-based methods, according to the CRPSS (which definition is in Appendix \ref{app4}).  This improvement is however less significant than for other weather variables  (see \citet{taillardat2016calibrated}). 
This corroborates \citet{hemri2014trends}'s conclusion that rainfall amounts are tricky to calibrate.
If the analogs method looks less performant, that might be imputable to the data depth of only 4 years.
 Indeed, this non-parametric technique is data-driven (such as QRF and GF) and needs more data to be effective (see e.g. \citet{van1994searching}).

\begin{figure}[!h]
  \includegraphics[width=\textwidth,angle=0]{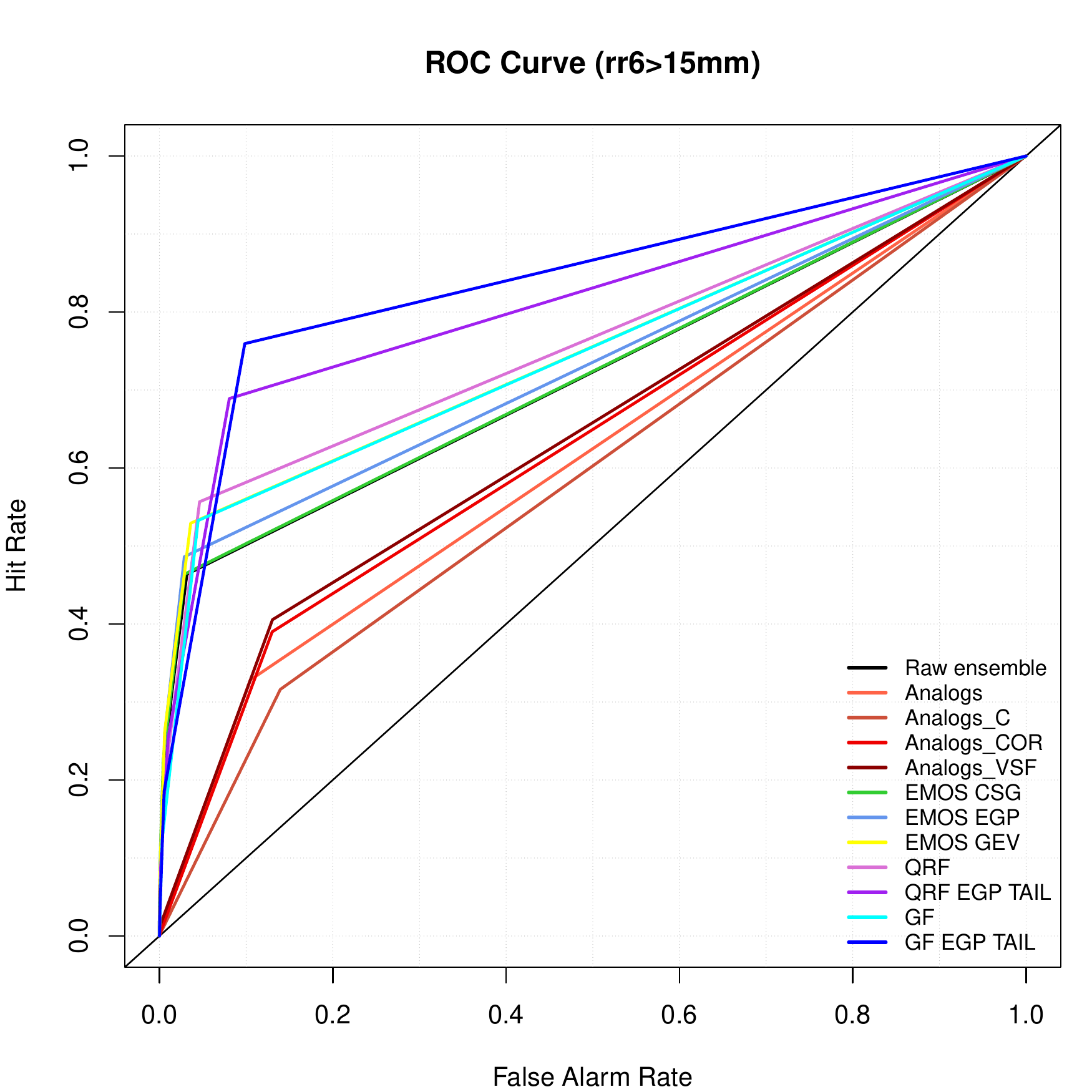}\\
  \caption{ROC Curves for the event of rain above 15mm. A ``good'' prediction must maximize hit rate and minimize false alarms. The analogs method lacks resolution. Tail extension methods show their gain in 
  a binary decision context.}\label{roc15}
\end{figure}

Concerning extreme events, Figure \ref{roc15}  shows the  benefit of the tail extension for forest-based methods. 
Note that we prefer to pay attention to the {\it value} of a forecast more than to its {\it quality}. According to \citet{murphy1993good}, the {\it value} can be defined as the ability of the forecast to help users to take better decisions.
 The {\it quality} of a forecast can be summarized by the area on the \textit{modelled} ROC curve (classically denoted by AUC), with
 some potential drawbacks exhibited by \citet{lobo2008auc,hand2009measuring}. \citet{zhu2002economic} made a link
 between optimal decision thresholds, value and cost/loss ratios.
 In particular, they show that the value of a forecast is maximized for the ``climatological''
 threshold and equals the hit rate minus the false alarm 
 rate which is the maximum of the Peirce Skill Score \citep{manzato2007note}. 
 This value corresponds to the upper left corner of ROC curves, which is of main interest in terms of
 extremes verification, as explained in  Section \ref{sec: zoom}. Several features already seen on  Figure~\ref{roc} can be observed on Figure~\ref{roc15}:
 analogs lack resolution  and the other post-processed methods compete more or less favourably with the raw ensemble. 
 Nonetheless, the other post-processing techniques stay better than the raw ensemble even for methods that cannot extrapolate observed values such as QRF and GF. 
 Note that QRF is rather surprisingly better than EMOS techniques.  Tail extension methods show their gain in a binary decision context.

\section{Discussion}\label{sec: end}
%

Throughout this study, we see that forest-based techniques  compete favourably with EMOS techniques. It is a good point to see that QRF and GF compared to EMOS exhibit nearly the
same kind of improvement  when focusing on rainfall amounts or on temperature and wind speed (see \citet{taillardat2016calibrated} Figures~6 and 13). It could be interesting to check these methods (especially GF) on smoother variables. 

Tail extension of these non-parametric techniques generates ensembles more tailored for extremes catchment.
However, reliability as well as resolution remain quite stable when extending the tail, so that our paradigm about verification (good extreme discrimination subject to satisfying overall performance) remains. 

One of the advantages of distribution-free calibration (analogs, QRF and GF) is that there is no assumption on the parameters to calibrate. This benefit is emphasized for rainfall amounts for which EMOS techniques have to be studied using different distributions. In this sense, the recent mixing method of \citet{baran2016mixture} looks  appealing.  A brand new alternative solution consists in working with (standardized) anomalies as done in \citet{dabernig2016spatial}.

Another positive aspect of the forest-based methods is that there is no need of a predictor selection. 
Concerning the analogs method, our results suggest that the work of \citet{genuer2010variable} 
could be a cheaper alternative to brute force algorithms like in \citet{keller2017statistical} for the weightning of predictors.
For analogs techniques, we can notice that the complete set of predictors gives the best results. In contrast, the choice of the set of predictors is still an ongoing issue for EMOS techniques regarding precipitation. For 
easier variables to calibrate, \citet{messner2017nonhomogeneous} shows that some variable selection can be effective.

 
 The tail extension can be viewed as a semi-parametric technique where the result of forest-based methods is used to fit a distribution. This kind of procedure can be  connected to the work of \citet{junk2015analog} who uses  analogs on EMOS inputs. An  interesting prospect  would be to bring forest-based methods in this context.
 
 
 A natural perspective  regarding spatial calibration and trajectory recovery could be to make use of block regression techniques as done in \citet{zamo2016improved}, or of ensemble copula coupling, as suggested by  \citep{bremnes2007improved, schefzik2016combining}.

Finally,  it appears that more and more weather services work on merging different forecasts from different sources (multi-model ensembles). In this context, an attractive procedure could be to combine raw ensembles and different methods of post-processing via sequential aggregation \citep{mallet2010ensemble, thorey2016online}, in order to get the best forecast according to the weather situations.

 \section*{Acknowledgments}
 Part of the work of P. Naveau has been supported by the ANR-DADA, LEFE-INSU-Multirisk, AMERISKA, A2C2, CHAVANA and Extremoscope projects. This work has been supported by Energy oriented Centre of
 Excellence (EoCoE), grant agreement number 676629, funded within the Horizon2020 framework of the European Union. This work has been supported by the LABEX MILYON (ANR-10-LABX-0070) of Universit\'e de Lyon, 
 within the program "Investissements d'Avenir" (ANR-11-IDEX-0007) operated by the French National Research Agency (ANR). Thanks to Julie Tibshirani, Susan Athey and Stefan Wager for providing gradient-forest source package.
\section*{Funding information}
LABEX MILYON, Investissements d'Avenir and DADA, ANR, Grant/Award Numbers: ANR-10-LABX-0070, ANR-11-IDEX-0007 and  ANR-13-JS06-0007; EoCoE, Horizon2020, Grant/Award Number: 676629; A2C2, ERC, Grant/Award Number: 338965

 \bibliographystyle{rss}

 \bibliography{references}
 
 \appendix

\section{Analogs method}\label{sec: analogs}
Contrary to EMOS, this technique is data-driven. An analog for a given location and forecast lead time is defined as a past prediction, from the same model, that has similar
values for selected features of the current model forecast. The method of analogs  consists in finding these closest past forecasts according to a given metric of the predictors' 
space to build an analog-based ensemble (see e.g. \citet{hamill2006probabilistic}). We assume here that close forecasts leads to close observations. Making use of analogs requires to choose both the set of predictors and the metric.
Concerning the metric, several have been tried like the Euclidean or the Mahalanobis distance but they have been outperformed by the   metric
provided in \citet{delle2013probabilistic}: 
 \begin{equation}
    \sum_{j=1}^{N_v} \frac{w_j}{\sigma_{f_j}}\sqrt{\sum_{i=-\tilde{t}}^{\tilde{t}}{\left(F_{j,t+i}-A_{j,t'+i}\right)}^2},
\end{equation}
where $F_t$ represents the current forecast at time $t$ for a given location. The analog for another time $t'$ at this same location is $A_{t'}$. The number of predictors is $N_v$ and $\tilde{t}$ is half the time window
used to search analogs. We standardize the distance by the standard deviation of each predictor $\sigma_{f_j}$ 
calculated on the learning sample for the considered location. In this study we take $\tilde{t}=1$ so the time window is $\pm 24$ hours the forecast to calibrate.
This distance has the advantages of being flow-dependent and thus defines a real weather regime associated with the research of the analogs. Note that one could weight the different predictors $f_j$ with $w_j$ and we fixed $w_j=1$ for all predictors in a first method (Analogs).
We have also tried two other weightning techniques using the absolute value of correlation coefficient between predictors and
 the response variable (Analogs\_COR) like in \citet{zhou2016new}, and a weightning based on the frequency of predictors' occurrences in 
 variable selection algorithm described in Appendix \ref{app3} (Analogs\_VSF). 
 Note finally that other weightning techniques have been considered \citep{horton2017global, keller2017statistical} but we did not use
 them in this study because of their computational cost.


\section{CRPS formula for EGP}\label{app1}

The CRPS for the distribution $F$ detailed in \ref{eq: EGPD} is:
\begin{eqnarray*}
 CRPS(F,y)&=&y(2F(y)-1) +\frac{\sigma}{\xi}(4\pi-2F(y)-\pi^2-1)\\
 && +\frac{2\kappa\sigma(1-\pi)}{\xi}\left[B\left({\left[1+\frac{\xi y}{\sigma}\right]}^{-\frac{1}{\xi}};1-\xi,\kappa\right)-(1-\pi)B(1-\xi,2\kappa)-\pi B(1-\xi,\kappa)   \right] ,
\end{eqnarray*}
where $0<\xi<1$ and $B(\,;\,,\,)$ and $B(\,,\,)$ denote respectively the incomplete beta and the beta functions.

\section{Variable selection using random forests}\label{app3}

We have seen that most parameters in EMOS and the distance used in analogs can be inferred using different sets of predictors. Contrary to the QRF and GF methods where the add of a useless
predictor does not impact the predictive performance (since
this predictor is never retained in the splitting rule), it can be misguiding for EMOS and analogs. We have therefore investigated some methods that keep the most informative meteorological variables and guarantee the best
predictive performance. Our first choice was to use the well-known Akaike information criterion and the Bayesian information criterion \citep{akaike1998information,schwarz1978estimating} but it resulted that the 
selection was not enough discriminant (too
many predictors kept in our initial set). The algorithm of \citet{genuer2010variable} has then been considered. Such an algorithm is appealing since it uses random forests (and we already have these objects from the QRF method) and it permits to
keep predictors without redundancy of information. For example this algorithm eliminates correlated predictors even if they are informative. A reduced set of predictors (mostly 3 or 4) is thus obtained, which avoids
misestimation
generated by multicolinearity. The method of variable selection used here is one among plenty others. The interested reader in variable selection using random
forests can refer to \citet{genuer2010variable} for detailed explanations.

\begin{figure}[!h]
  \includegraphics[width=\textwidth,angle=0]{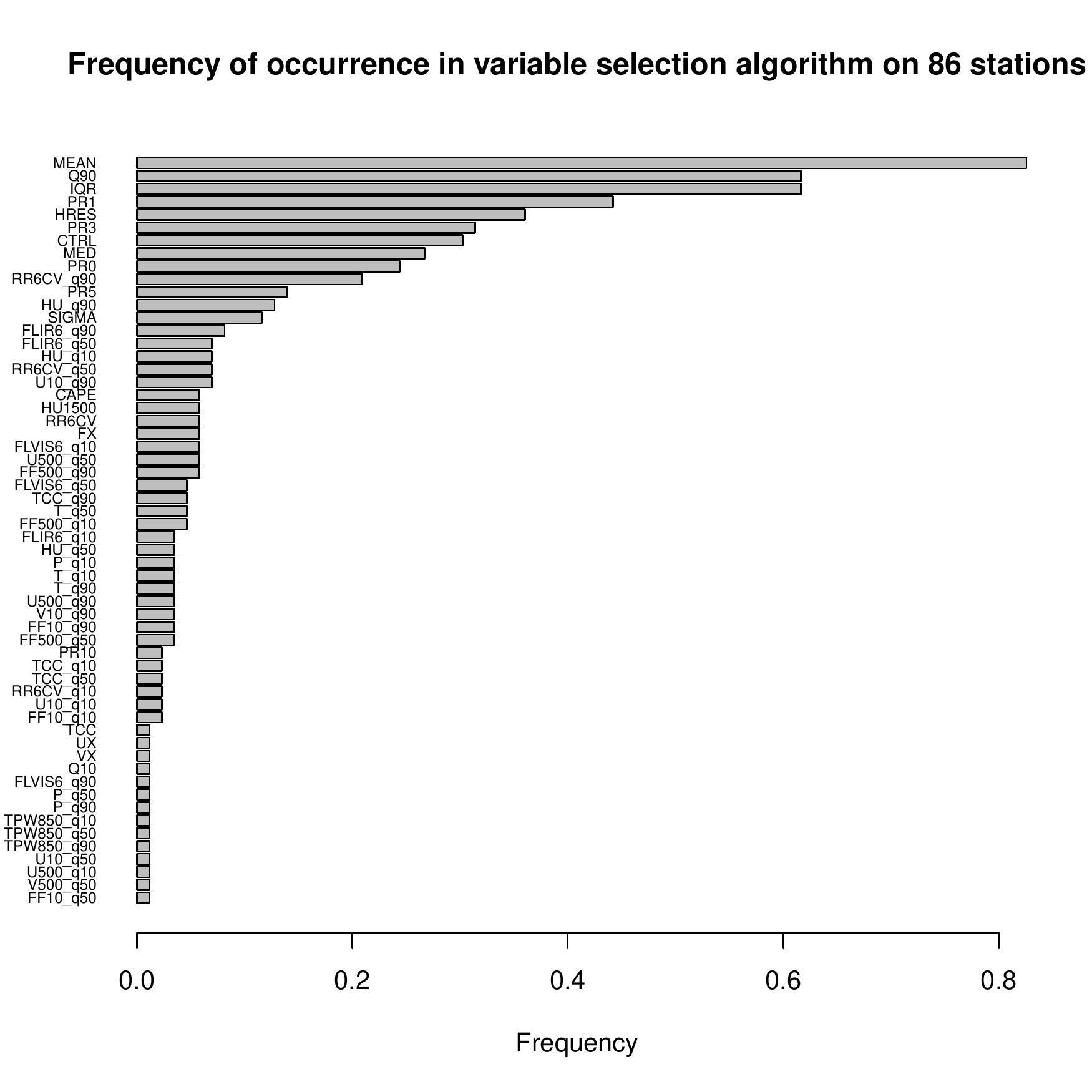}\\
  \caption{Frequency of predictors' occurrence in variable selection algorithm. Variables representing central and extreme tendencies are preferred. Some covariables like CAPE, FX or HU can be retained. It is 
  interesting to see that only one third of the predictors of the set is taken more than in $10\%$ of the cases.}\label{procc}
\end{figure}

The variable selection algorithm is used to keep the first predictors (max 4 of them) that form the set of predictors for each location. Figure \ref{procc} shows the ranked frequency of each
chosen predictor. Predictors never retained are not on this figure. We can see here that only one third of the predictors in A are retained at least in $10\%$ of the cases. Moreover, predictors representing central and extreme tendencies are preferred. 
Some predictors appear that differ from rainfall amounts ; see CAPE, FX or HU. It is not surprising since these parameters are correlated with storms. It is not shown here but when the MEAN variable is not selected, 
either MED or CTRL stands in the set. This shows that the algorithm mostly selects just one information concerning central tendency and avoid potential correlations. So the results concerning the variable algorithm selection 
seem to be sound. Last but not least, one notices that the predictors of the set C are often chosen. This remark confirms both the robustness of the algorithm and the relevance of previous studies on precipitation concerning the choice
 of the predictors.


\section{Verification of ensembles}\label{app4}

We recall here some facts about the scores used in this study.

 
 \subsection{Reliability}
 
 Reliability between observations and a predictive distribution can be checked by calculating $Z'=F(Y)$ where $Y$ is the observation and $F$ the cdf of the associated
predictive distribution. Subject to calibration, the random variable $Z'$ has a standard uniform distribution \citep{gneiting2014probabilistic} and we can check 
ensemble bias by comparing $\mathbf{E}(Z')$ to $\frac{1}{2}$ and ensemble dispersion by comparing the variance $\mathrm{Var}(Z')$ to $\frac{1}{12}$. This approach is applied to a $(K+1)$ ranked ensemble forecast
 using the discrete random variable $Z=\frac{\mathrm{rank}(y)-1}{K}$. Subject to calibration, $Z$ has a discrete standard uniform distribution with $\mathbf{E}(Z)=\frac{1}{2}$
 and a normalized variance $\mathbf{V}(Z)=12\frac{K}{K+2}\mathrm{Var}(Z)=1$.\\

Another tool used to assess calibration is the entropy: \begin{equation*}\Omega=\frac{-1}{\log(K+1)}\sum_{i=1}^{K+1} f_i\log(f_i).\end{equation*}
 For a calibrated system the entropy is maximum and equals $1$. \citet{tribus2013rational} showed that the entropy is an indicator of reliability linked to the Bayesian psi-test. It is also a proper measure of reliability used in the divergence score described in \citet{weijs2010kullback, roulston2002evaluating}.
 
 These quantities are closely related to rank histograms which are discrete version of Probability Integral Transform (PIT) histograms. However if one can assume the property of flatness of these histograms, \citet{jolliffe2008evaluating} exhibit a test accounting for the slope and the shape of rank
  histograms. In a recent work, \citet{zamtest} extends this idea for accounting the presence of wave in histograms as seen in \citet{scheuerer2015statistical, taillardat2016calibrated}. A more
   complete test can thus be implemented that tests each histogram for flatness. Such a test is called the JPZ test (for Jolliffe-Primo-Zamo). The results of the JPZ test is provided for each method in the  \ref{app5}.
 
 \subsection{Scoring rules}

Following \citet{gneiting2007probabilistic,gneiting2007strictly,brocker2007scoring}, scoring rules assign numerical scores to probabilistic forecasts
and form attractive summary measures of predictive performance, since they address calibration and sharpness simultaneously. These scores are generally negatively oriented 
and we wish to minimize them. A \textit{proper} scoring rule is designed such that the expected value of the score is minimized by the perfect forecast, ie. when the observation is drawn from
the same distribution than the predictive distribution. The \textit{Continuous Ranked Probability Score} (CRPS) \citep{matheson1976scoring,hersbach2000decomposition} is defined directly
in terms of the predictive cdf, $F$, as:
\begin{equation*}CRPS(F,y)=\int_{-\infty}^{\infty} (F(x) - \mathbf{1}\{x\geq y\})^2 \, \mathrm{d}x .\end{equation*}

Another representation \citep{gneiting2007strictly} shows that:
\begin{equation*}CRPS(F,y)=\mathbf{E}_F|X-y|-\frac{1}{2}\mathbf{E}_F|X-X'| ,\end{equation*}
where $X$ and $X'$ are independent copies of a random variable with distribution $F$ and finite first moment.

An alternative representation for continuous distributions using L-moments \citep{hosking1989some} is:
\begin{equation*}CRPS(F,y)=\mathbf{E}_F|X-y|+\mathbf{E}_F(X)-2\mathbf{E}_F(XF(X)).\end{equation*}

Throughout our study, if $F$ is represented by an ensemble forecast with K members $x_1,\ldots,x_K \in \mathbf{R}$, we use a so-called fair estimator of the CRPS \citep{ferro2014fair} given by :
\begin{equation*}
 \widehat{CRPS}(F,y)=\frac{1}{K}\sum_{i=1}^K |x_i-y| - \frac{1}{2K(K-1)}\sum_{i=1}^{K}\sum_{j=1}^{K} |x_i-x_j| .\end{equation*}

 Notice that all CRPS have been computed following the recommendations of the Chapter 3 in \citet{zamtest}.

 We can also define the skill score in term of CRPS between an ensemble prediction system A and a baseline B, in order to compare them directly:

\begin{equation*}CRPSS(A,B)=1-\frac{CRPS_A}{CRPS_B}\end{equation*}
The value of the CRPSS will be positive if and only if the system A is better than B for the CRPS scoring rule.

\section{Rank histograms boxplots}\label{app5}

\begin{figure}[!h]
  \includegraphics[width=\textwidth,angle=0]{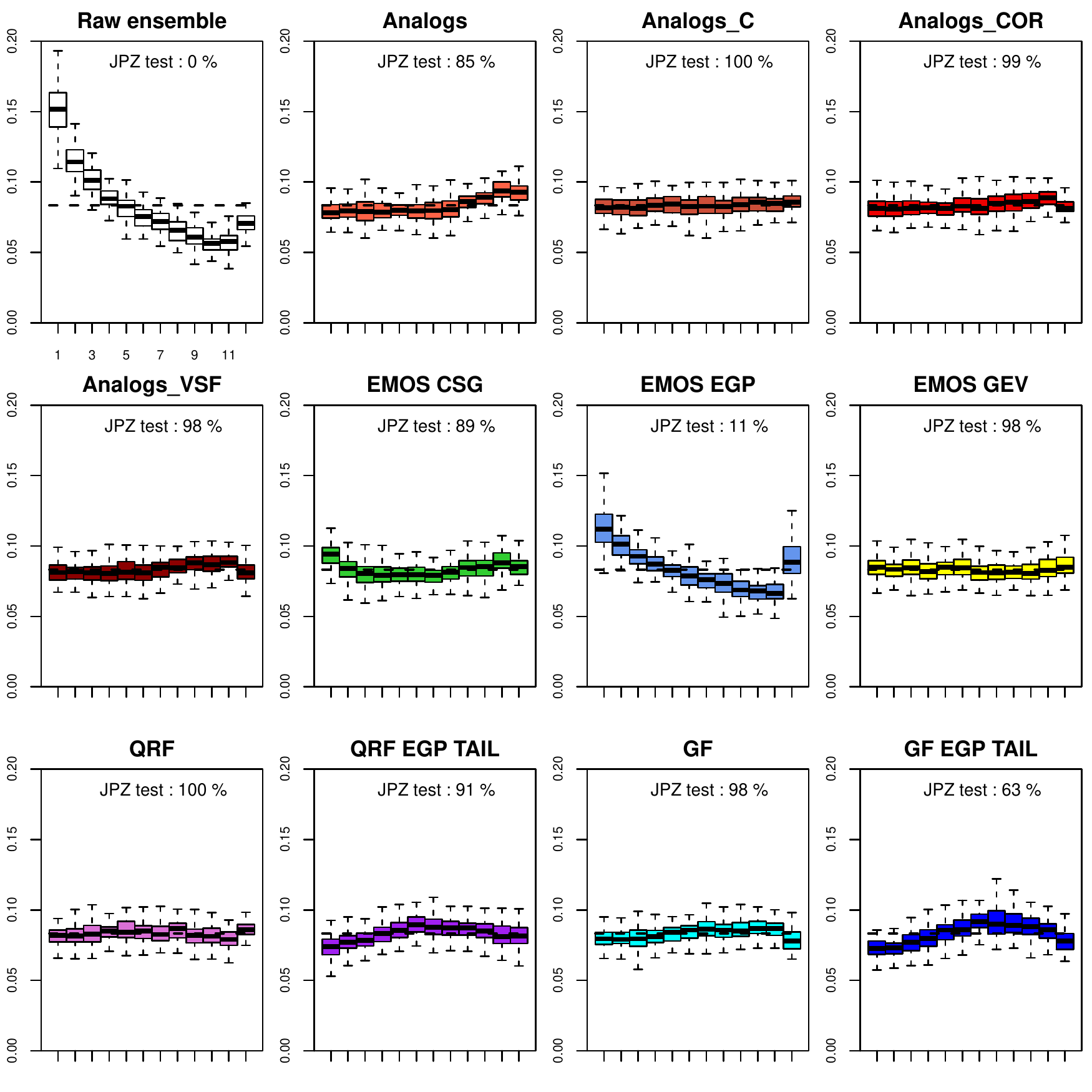}\\
  \caption{Boxplots of rank histograms for each technique according to the locations. The proportion of rank histograms for which the JPZ test does not reject the flatness hypothesis is also provided. The results confirm the Table \ref{res}.}\label{boxhdr}
\end{figure}

\end{document}